\documentclass[lettersize,journal,letterpaper,compact]{IEEEtran}
\usepackage{amsmath,amsfonts}
\usepackage{algorithmic}
\usepackage{algorithm}
\usepackage{array}
\usepackage[caption=false,font=normalsize,labelfont=sf,textfont=sf]{subfig}
\usepackage{textcomp}
\usepackage{stfloats}
\usepackage{url}
\usepackage{verbatim}
\usepackage{graphicx}
\usepackage{cite}

\usepackage{times}
\usepackage{soul}
\usepackage{url}
\usepackage[hidelinks]{hyperref}
\usepackage[utf8]{inputenc}
\usepackage[small]{caption}
\usepackage{graphicx}
\usepackage{amsmath}
\usepackage{amsthm}
\usepackage{booktabs}
\usepackage{algorithm}
\usepackage{algorithmic}
\usepackage[switch]{lineno}
\usepackage{amssymb}
\usepackage{xcolor}
\usepackage{tabularx}
\newcolumntype{C}{>{\centering\arraybackslash}X}
\newcolumntype{L}{>{\raggedright\arraybackslash}X}
\begin{document}

\title{Group-wise Supervision with Focal-Dice Loss for Long-Tailed Indoor Semantic Occupancy Prediction}

\author{Qi Zheng, Zihuang Su, and Xiao Pan
\thanks{Qi Zheng, Zihuang Su, and Xiao Pan are with the Guangdong Key Laboratory of Intelligent Information Processing, College of Electronics and Information Engineering, Shenzhen University, Shenzhen 518060, China
        {\tt\small \{qiz\}@szu.edu.cn}}%
}



\maketitle

\begin{abstract}
 Recently, 3D semantic occupancy prediction has garnered increasing attention for understanding the indoor scene. However, unlike structured outdoor environments, indoor scenes feature a high diversity of object categories that exhibit a severe long-tailed distribution, which has become a core bottleneck limiting the performance of existing models. To tackle this challenge, we propose a novel method, Group-UFD Occ, based on hierarchical semantic supervision and synergistic loss optimization. At the architectural level, we introduce a fine-grained semantic grouping strategy and design multi-scale, parallel ``main-expert'' prediction heads to guide the model in efficiently learning tail-class features through deep regularization. At the optimization level, we introduce the Unified Focal-Dice (UFD) loss. This synergistic loss function dynamically focuses on hard samples at the per-voxel level. Meanwhile, it simultaneously optimizes the geometric integrity of predicted objects from a region-based perspective. We conducted experiments on the large-scale EmbodiedScan dataset. The results demonstrate that our method yields
 a relative improvement of 11.38\% over the baseline, with substantial accuracy gains in several critical long-tailed categories.
\end{abstract}


\section{Introduction}
\IEEEPARstart{A}{ccurate} 3D world perception is fundamental for advanced artificial intelligence applications such as embodied AI~\cite{liu2025aligning,11123690}, augmented reality~\cite{lynam2025augmented} and autonomous driving~\cite{10740797}. Different from 3D object detection~\cite{guan2024gramo,10325591,11123741} that primarily describes the world by predicting sparse 3D bounding boxes for objects, semantic occupancy prediction has emerged as a novel paradigm of 3D perception. It discretizes the scene into a dense voxel grid and predicts the occupancy state and semantic class for each voxel, thereby enabling a unified representation of arbitrarily shaped objects.

Although semantic occupancy prediction has achieved significant progress in autonomous driving in outdoor environments~\cite{ma2024cotr,tang2024sparseocc,xu2025survey,zhang2026vision}, its adaptation to indoor scenes remains challenging. Among the challenges, the long-tailed distribution of object categories stands out as the core bottleneck that limits the performance of current indoor semantic occupancy prediction models. Large-scale indoor datasets like EmbodiedScan~\cite{wang2024embodiedscan,dai2017scannet,wald2019rio,chang2017matterport3d,silberman2012indoor} cover more than 80 object classes, where the sample count (voxel) of a few head classes (e.g. wall, floor) is hundreds or thousands of times higher than that of numerous tail classes.
Under such severe long-tailed class distribution, the gradient signals in conventional deep learning models can be almost entirely dominated by the head classes during optimization, while features for the more critical tail classes are consequently ignored. This severely undermines the model's generalization capability and curtails its practical utility in real-world environments.


While recent works like COTR~\cite{ma2024cotr} propose coarse-to-fine grouping for outdoor scenes with limited classes, we found such strategies less effective indoors: the dramatically larger taxonomy (e.g., 80+ classes) leads to inefficient grouping hierarchies, while subtle inter-class distinctions weaken feature discriminability. To overcome these limitations, we propose a fine-grained semantic grouping strategy that reorganizes categories based on indoor-specific attribute similarities. To further address indoor long-tail challenges, we draw inspiration from medical image segmentation~\cite{lin2017focal,milletari2016v}, where tailored loss functions like Focal and Dice losses effectively handle class imbalance. This motivates our design of a unified loss that adapts similar principles to 3D occupancy prediction.

In this paper, we propose a novel optimization framework named Group-UFD Occ, which introduces synergistic innovations at both the architecture design and loss function levels. At the architectural level, we formulate a Hierarchical Semantic Grouping strategy. This strategy restructures all classes into multiple semantically cohesive subsets and equips them with parallel ``main-expert" prediction heads, guiding the model to learn discriminative features for rare classes via deep regularization. At the optimization level, we devise a Unified Focal-Dice (UFD) Loss. It leverages Focal loss to dynamically focus on hard-to-recognize rare categories in a per-voxel manner, thereby enhancing the gradient signals for tail classes. Concurrently, it capitalizes on the global region-based structure optimization capability of Dice loss. This loss encourages the model to improve the geometric integrity of its predictions by directly maximizing the overlap between predicted objects and their ground truth. By utilizing these two core loss functions, our method improves the recognition accuracy for tail categories from two complementary perspectives: focusing on rare classes and optimizing region-based segmentation. The results demonstrate that our method achieves significant performance gains.

Our main contributions are as follows:
\begin{itemize}
    \item We propose a novel semantic grouping strategy that reorganizes the dataset's classes based on two principles: semantic cohesion and frequency mixing. This strategy enhances the model's semantic discriminability for tail classes through hierarchical supervision.
    \item We devise a Unified Focal-Dice loss for optimization. It can dynamically focus on rare classes while balancing the precision and recall of the occupancy predictions, thereby improving recognition accuracy.
    \item Experiments on the EmbodiedScan dataset demonstrate that our method, Group-UFD Occ, not only achieves good performance on head classes but, more importantly, improves the model's recognition capability on critical tail classes, leading to a gain in the overall mIoU.
\end{itemize}

\section{Related Work}
\subsection{3D Semantic Scene Completion}
3D Semantic Scene Completion (SSC) is generally considered the precursor to the task of indoor semantic occupancy prediction. Its goal is to infer the complete 3D spatial structure and corresponding semantics from a single-view, occluded RGB or RGB-D image. SSCNet~\cite{song2017semantic}, as the foundational work in this field, was the first to formally define and propose an end-to-end solution for semantic scene completion. Subsequent research has primarily evolved along two paths. One is to directly optimize the efficiency of 3D processing by introducing more efficient network architectures, such as sparse 3D convolutions. The other is to more fully leverage the rich texture and semantic information from RGB images, guiding the completion of the 3D scene by designing sophisticated 2D-to-3D feature fusion mechanisms~\cite{liu2021lmscnet}. MonoScene~\cite{cao2022monoscene} was the first to successfully achieve high-quality indoor SSC using only monocular RGB images, removing the dependency on depth sensors and proving the feasibility of building pure vision-based 3D perception algorithms.

\begin{figure*}[th]  
    \centering
    \setlength{\abovecaptionskip}{3pt}
    \setlength{\belowcaptionskip}{3pt}
    \includegraphics[width=1.0\textwidth, keepaspectratio]{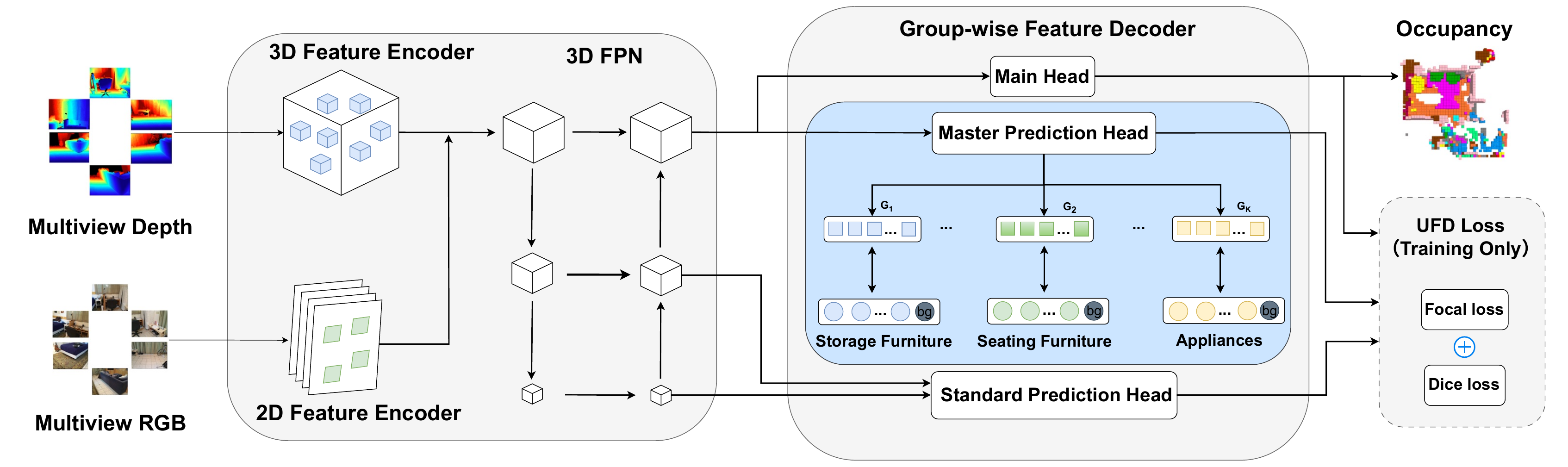}
    \caption{The overall framework of our Group-UFD Occ. The 2D and 3D encoders first extract image features and sparse point cloud features from the multi-view RGB images and depth maps, respectively. After concatenation, the two types of features are fused by a 3D FPN, which then outputs voxel grid features at different scales. The group-wise feature decoder receives these multi-scale voxel features and leverages the hierarchical semantic grouping strategy in concert with the UFD loss to enhance the model's long-tail discrimination capability.}  
    \label{f1}
\end{figure*}

\subsection{3D Semantic Occupancy Prediction}
\subsubsection{Occupancy Prediction based on Explicit Representation}
The core idea of methods based on explicit representation is to discretize the scene into a voxel grid and employ an encoder-decoder network to predict the occupancy state and semantic class for each voxel. In the autonomous driving domain, early approaches relied on LiDAR-based point cloud inputs. The BEV-based paradigm~\cite{10791908} pioneered the exploration of vision-only occupancy prediction~\cite{tong2023scene} by transforming and aggregating multi-view image features onto a unified Bird's-Eye-View (BEV) plane. TPV (Tri-Perspective View)~\cite{huang2023tri}, in contrast, projects features onto three mutually orthogonal planes. Both are exemplars of efficient 2.5D compressed representations of 3D features. Subsequent work has primarily evolved in two directions. The first is toward more efficient 2D-to-3D feature lifting mechanisms~\cite{philion2020lift}, which aim to map 2D features into 3D space more accurately and effectively. The second direction involves leveraging advanced network architectures
to explore more effective fusion of temporal information or features from multi-modal sensors~\cite{ma2024cotr,ming2024occfusion,TIAN2026131305,wang2023openoccupancy}.

In the context of indoor scenes, 
ISO utilizes a pre-trained depth prediction network to achieve accurate depth estimation, providing a new benchmark for monocular vision-based occupancy prediction in indoor environments~\cite{yu2024monocular}. SliceOcc ingeniously proposes a paired-slice representation along the vertical axis~\cite{li2025sliceocc}. EmbodiedScan~\cite{wang2024embodiedscan} addresses ego-centric 3D scene understanding by introducing a large-scale multimodal dataset and a baseline model.

\subsubsection{Occupancy Prediction based on Implicit Representation}
Methods based on implicit representation take a different approach, foregoing a predefined discrete scene structure in favor of encoding the entire 3D scene with a continuous function or a set of primitives. 
For example, RenderOcc~\cite{pan2024renderocc} innovatively projects the predicted 3D semantic occupancy field back into 2D semantic segmentation maps via a differentiable volume rendering mechanism. This allows for dense supervision by comparing against 2D ground truth, circumventing the need for 3D annotations.
EmbodiedOcc~\cite{wu2025embodiedocc} explores how 3DGS can be utilized to build a dynamic and continuously updated environmental model for an embodied agent, thereby supporting more efficient navigation and interaction.

However, despite their architectural differences, the aforementioned methods all inherit a common bottleneck from real-world data: the long-tailed distribution of semantic categories. This issue significantly impairs performance on rare but critical objects. Addressing this challenge and improving the model's learning capability on tail categories is the central motivation for our work.

\section{Method}
We follow the general setting in EmbodiedScan~\cite{wang2024embodiedscan}. The semantic occupancy prediction task takes multi-view RGB and depth images as input, and predicting the occupancy state and semantic category for every voxel in the scene, resulting in an output volume $O \in \mathbb{R}^{C \times X \times Y \times Z}$.
Here, $X, ~Y, ~Z$ represent the spatial dimensions of the voxel grid, and $C$ is the number of class labels, where $C=0$ denotes an empty state.

\subsection{Overall Architecture}
In this paper, we propose a novel semantic occupancy prediction method for indoor scenes, targeting the prevalent long-tailed distribution problem of object categories.
The overall framework of our proposed method, named Group-UFD Occ, is illustrated in Figure~\ref{f1}. 
Our framework primarily consists of three components: 1) 2D and 3D feature encoders for extracting image and point cloud features; 2) a group-wise feature decoder that enhances tail-category recognition through group-wise supervision; and 3) the Unified Focal-Dice loss, which is designed to further optimize the generation of geometric structures and the gradient signals for rare classes.

We adopt the same modules for feature encoding as in EmbodiedScan~\cite{wang2024embodiedscan}. The feature encoder first employs a pre-trained ResNet-50 and a MinkResNet to extract 2D image features and sparse 3D point features from the multi-view RGB and depth inputs, respectively. The 2D features are then lifted into the 3D space. After densifying the sparse 3D features, they are concatenated with the projected 2D features. This fused feature volume is subsequently processed by a 3D Feature Pyramid Network (3D FPN), which further refines and fuses the features to produce voxel representations at three different resolutions—high, medium, and low—for use by the subsequent decoder.

\subsection{Hierarchical Semantic Grouping Strategy}

When re-grouping categories in a dataset, there are typically two common criteria: the semantics of the category object and the number of samples in that category (i.e., the number of voxels contained in that category across all scenarios in the dataset). For example, COTR adopts a coarse-to-fine grouping scheme that categorizes classes into foreground, background, and empty regions, followed by subdividing them based on similar sample size. Such a grouping scheme has certain effects in outdoor scenes, but in diverse and complex indoor scenes, its effectiveness is quite limited. Indoor object categories cannot be naively split into purely foreground or background categories. Relying solely on sample-size similarity easily groups semantically divergent categories together or causes huge inter-group variance. Moreover, in indoor environments, the difference in sample size between the head category and the long tail category is often larger compared to outdoor scenes. This grouping method can cause the gradient of groups with sparse sample sizes to be overwhelmed.

Therefore, 
we design an LLM-assisted semi-automated grouping strategy, based on the principles of semantic cohesion and selective frequency mixing. It works as follows.
\paragraph{Initial Frequency-Aware Clustering} We first input all 80 semantic categories from the EmbodiedScan dataset along with their voxel appearance frequencies (sample sizes) into Gemini 2.5 Pro. We instructed the LLM to group these classes according to semantic relationships and frequency-mixing principles without specifying the target number of groups. This yielded an initial set of 6 semantic groups, each naturally balancing high-, medium-, and low-frequency categories to facilitate intra-group knowledge transfer.

\paragraph{Semantic Constrained Multi-Granularity Expansion} Next, we enforced strict hierarchical semantic cohesion rules onto the LLM: semantic grouping priority was explicitly restricted in descending order of functional utility, 3D geometric shape, and co-occurring scene context. Unclassifiable or extremely ambiguous categories were assigned to a ``miscellaneous'' fallback group. Under these structural constraints, the LLM reshaped and merged the initial clusters into multi-granularity grouping candidate schemes (specifically, configurations with 3, 7, 12, and 17 groups). A minor manual review was conducted to rectify obvious semantic mismatches in boundary categories.

Given that each category can only belong to one group, we divide all categories into multiple expert groups.
For semantic categories and empty categories that are not within any group, they are collectively considered as the ``others" category. 
The empirical validation performance confirmed that the 12-group configuration achieves the optimal balance between expert capacity and local representation power.

\subsection{Semantic Group-wise Feature Decoder}
To enable effective learning over the long-tailed distribution, we designed a novel feature decoder built upon a multi-Task, multi-Scale supervision framework, which is applied at the output of the 3D FPN at three different resolution scales: high, medium, and low. This decoder is not a monolithic module but rather a parallel system where a Main Prediction Head, Grouped Expert Heads, and Standard Prediction Heads work in synergy. The prediction heads corresponding to the outputs of different voxel resolution scales are as follows.

\paragraph{Prediction Head at High Resolution Scale}
At the high-resolution feature map output by the 3D FPN, we deploy two types of prediction heads in parallel, a main head and M grouped expert heads. The main head is a standard classifier that performs an 81-way classification. It serves as the exclusive decision-maker during the inference phase, generating the final semantic occupancy prediction after receiving the optimally regularized features. The Grouped Expert Heads are a cluster composed of $M$ independent, smaller prediction heads. Each expert head is responsible for a designated semantic subset (e.g., seating furniture, appliances). Specifically, the $j$-th expert head ($j \in \{1, 2, \dots, M\}$) oversees $m_j + 1$ categories, outputting $m_j + 1$ class prediction probabilities, where $m_j$ denotes the number of semantic categories belonging to the $j$-th group. Each grouped expert head is uniquely dedicated to supervising andpredicting its designated subset of target categories. To isolate the minority features from mass environmental noise, the \texttt{empty} space and all other 
out-of-group foreground categories are intentionally coalesced into a single 
residual class denoted as ``others'' (labeled as \textit{background} in Figure~\ref{f1}). 

\paragraph{Prediction Head at Medium and Low Resolution Scale}
For the medium and low resolution feature maps output by the 3D FPN, we equip each with a standard 81-way classification head, which has the same network structure as the main head. They provide deep supervision by creating more direct gradient backpropagation paths to the shallower layers of the network. This not only alleviates the vanishing gradient problem and accelerates model convergence but also encourages the model to learn the overall scene layout and coarse geometric structures at more abstract levels.

We use the same loss function in all three prediction heads, and in order to improve the effectiveness of hierarchical semantic grouping supervision, we set the weight of the loss values of the main head and group prediction heads to 1:1. For standard prediction heads with medium and low resolution scales, we multiply them by a weight that decreases gradually with each resolution. 

\subsection{UFD Loss for Long-Tailed Prediction}
We begin by analyzing the optimization objectives of the EmbodiedScan baseline. Its loss function is a composite of a standard multi-class Cross-Entropy (CE) loss and two auxiliary terms: a semantic affinity loss (sem-scal) and a geometric affinity loss (geo-scal). Our analysis identifies three fundamental flaws in this design. First, its standard Cross-Entropy loss is overwhelmed by gradients from high-frequency head classes, causing the model to neglect rare tail classes. Furthermore, the auxiliary sem-scal and geo-scal losses are inefficient, as they optimize for indirect proxy metrics and divert learning resources to identifying empty space rather than foreground objects. Finally, the entire framework lacks a direct, region-based objective to enforce the shape integrity and boundary fidelity of predicted objects. To overcome these limitations, we replace the baseline's loss system with our Unified Focal-Dice (UFD) Loss, which combines Focal loss and Dice loss to simultaneously address class imbalance and improve segmentation quality.

First, our overall task training objective $\mathcal{L}_{\text{total}}$ is as follows:
\begin{equation}
    \mathcal{L}_{\text{total}} = \sum_{i=0}^{2} \mathcal{L}_{\text{scale}\_i},
\end{equation}
where $i \in \{0, 1, 2\}$ represents voxels of three different resolution scales from high to low. For any scale $i$, let $\Omega^{(i)}$ denote the valid voxel spatial domain filtered by the ignore index (\texttt{ignore\_index=255}), and let 
$N^{(i)} = |\Omega^{(i)}|$ represent the total number of valid samples. 
For any valid voxel $v \in \Omega^{(i)}$, its ground-truth category label 
is denoted as $y_v$.

\subsubsection{Loss for High Resolution Scale} 
We deploy a main prediction head alongside $M$ parallel grouped expert heads on the high-resolution features map output by the 3D FPN, where we set $M=12$ in our experiments. 
The loss term for the high-resolution scale ($i=0$) is
\begin{align}
    \mathcal{L}_{\text{scale}\_0} &= \frac{1}{N^{(0)}} \sum_{v \in \Omega^{(0)}} \ell_{\text{unified}}^{\text{main}}(v)  \notag \\ 
    &+ \lambda_{\text{group}} \cdot \left( \frac{1}{M} \sum_{j=1}^{M} \frac{1}{N^{(0)}} \sum_{v \in \Omega^{(0)}} \ell_{\text{unified}}^{\text{group}\_j}(v) \right).
\end{align}
The main head and each grouped expert head share the identical composite loss closure $\ell_{\text{unified}}$, consisting of the multi-class Focal loss and Dice loss. The unified loss of the expert heads is scaled by a grouping balance weight $\lambda_{\text{group}}$, which is set to $1.0$ in this work. Consequently, 
the main head and the grouped expert heads contribute equally to the scale-specific 
supervision with a 1:1 loss ratio. This formulation converts the complex multi-class problem into a highly focused, group specific detection task, forcing each expert to robustly discriminate its specialized long-tail categories against the entire rest of the scene landscape.

\subsubsection{Loss for Medium and Low Resolution Scale} 
For the lower resolution scales ($i \in \{1, 2\}$), the model retains only a single 
standard prediction head and applies a scale-dependent penalty weight $0.5^i$. 
\begin{equation}
    \mathcal{L}_{\text{scale}\_i} = 0.5^i \cdot \frac{1}{N^{(i)}} \sum_{v \in \Omega^{(i)}} \ell_{\text{unified}}^{\text{main}}(v), \quad i \in \{1,2\}.
\end{equation}
This exponentially decaying weight prevents coarse-grained gradients from 
interfering with the optimization of the main feature pyramid structure.
The focal loss and dice loss are added directly with a weight of 1:1, and the formula is
\begin{equation}
    \ell_{\text{unified}}(v) = \ell_{\text{focal}}(v) + \ell_{\text{dice}}(v).
\end{equation}

\begin{table*}[!htbp]
  \centering
  \resizebox{\linewidth}{!}{
  \begin{tabular}{l|c|c c c c c c c c c c c c c c c c}
    \toprule
    Method & mIoU(\%) & empty	& floor	& wall	& chair	& cabinet	& door	& table	& couch	& shelf	& window	& bed	& curtain	& refri.	& plant	& stairs & toilet	\\							
    \midrule
    
EmbodiedScan~\cite{wang2024embodiedscan} & 19.97 & 71.21 & 64.92 & 55.00 & 52.04 & 27.35 & 33.97 & \textbf{47.93} & 46.26 & 31.87 & 27.98 & 46.58 & 46.56 & 24.05 & 39.01 & 24.40 & \textbf{67.79}\\
    
EmbodiedScan$^*$~\cite{wang2024embodiedscan} & 20.39 & 73.62 & 69.40 &54.90 & 52.56 & 28.59 & 33.80 & 39.99 & 44.91 & 40.63 & 28.05 & 46.87 & 48.53 & 15.75 & 34.35 & 35.92 & 63.54\\

COTR$^\dagger$~\cite{ma2024cotr} & 20.31 & 73.81 & 70.34 & 53.99 & 53.72 & 29.74 & 31.44 & 41.34 & 47.80 & 42.96 & 25.70 & 50.96 & 45.96 & 17.52 & 37.70 & 31.13 & 62.68 \\

    \midrule
Group-UFD Occ (Ours) & \textbf{22.71} & \textbf{73.99} & \textbf{71.13} & \textbf{54.92} & \textbf{55.92} & \textbf{34.15} & \textbf{36.84} & 44.47 & \textbf{54.19} & \textbf{45.70} & \textbf{29.64} & \textbf{59.32} & \textbf{54.20} & \textbf{24.60} & \textbf{39.32} & \textbf{42.74} & 66.81\\

 \bottomrule
  \end{tabular}
  }
  \caption{Comparison of multi-view 3D occupancy prediction. Following EmbodiedScan, we report results on the 16 most common categories (out of 81 total). The best results are \textbf{bold}. $^*$: reproduced results; $^\dagger$: adapted to indoor scenes; ``refri.'': ``refrigerator''. }
  \label{tab:main_result}
\end{table*}

\begin{table*}[!htbp]
  \centering
  \resizebox{\linewidth}{!}{
  \begin{tabular}{l|c|c|c c c c c c c c c c c c c c c c}

    \toprule

    Method & mIoU(\%) &  Tail-mIoU(\%) & commode	& tv & monitor & radiator & shower & mat & ottoman & wm. & towel & column & wf. & rack & stove & pillar & blinds & bar\\
    
    \midrule

EmbodiedScan$^*$~\cite{wang2024embodiedscan}  & 20.39 & 13.28 & 0.99 & 21.82 & 52.13 & 38.51 & 1.06 & \textbf{2.00} & 14.57 & 17.95 & 22.56 & 2.84 & 0.13 & \textbf{0.50} & 17.29 & \textbf{5.21}	& \textbf{1.84} & 13.04 \\

COTR$^\dagger$~\cite{ma2024cotr}  & 20.31  & 13.86 & 2.95 & \textbf{24.81}	& 53.96	& 38.17	& 1.13 & 0.60	& 12.11	& 18.84	& 24.43	& \textbf{3.12}	& \textbf{0.26}	& 0.06	& 20.81	& 5.07	& 1.15	& 14.21 \\	

    \midrule
Group-UFD Occ (Ours) & \textbf{22.71}  & \textbf{17.12} & \textbf{4.29} & 24.12 & \textbf{54.57} & \textbf{43.63} & \textbf{3.03} & 0.18	& \textbf{19.43}	& \textbf{26.75}	& \textbf{28.58}	& 1.20	& 0.00	& 0.00	& \textbf{22.39}	& 0.00	& 0.00 & \textbf{45.83} \\

    \bottomrule
  \end{tabular}
  }
  \caption{Comparison of multi-view 3D occupancy prediction for tail categories (bottom 50\%). Tail-mIoU is the mean of long tail categories listed in the table. The best results are \textbf{bold}. $^*$: reproduced results; $^\dagger$: adapted to indoor scenes; ``wm.'': ``washing  machine'', ``wf.'': ``window frame''.}
  \label{tab:minor_result}
\end{table*}

\begin{table*}[htbp]
  \centering
  \resizebox{\linewidth}{!}{
  \begin{tabular}{l|c|c|c c c c c c c c c c c c c c c c}

    \toprule

    Method & mIoU(\%) & Tail-mIoU(\%) & carpet	& copier & drawer & piano & washbasin & countertop & oven & partition & microwave & printer & mailbox & bicycle & frame & ee. & beam & roof\\   
    \midrule
EmbodiedScan$^*$~\cite{wang2024embodiedscan} & 20.39 & 6.33 & 3.59
& 23.10 & \textbf{3.59} & 4.76 & 1.34 & \textbf{2.15} & \textbf{8.72} & \textbf{0.67} & 14.31 & 31.34 & 0.00 & 7.56 & 0.00 & \textbf{0.08} & 0.00 & 0.00\\
    \midrule
COTR$^\dagger$~\cite{ma2024cotr} & 20.38 & 5.82	& 0.00	& 29.55	& 0.80	& 6.01	& \textbf{1.58}	& 0.92	& 6.57	& 0.14	& 17.12 & 22.07	& 0.00	& \textbf{8.44}	& 0.00	& 0.00	& 0.00	& 0.00	\\																										
    \midrule
Group-UFD Occ (Ours) & \textbf{22.71}  & \textbf{7.92} & \textbf{5.37} & \textbf{37.16} & 0.00 & \textbf{9.85} & 0.00 & 0.47 & 1.96 & 0.00 & \textbf{20.50} & \textbf{46.03} & 0.00 & 5.36 & 0.00 & 0.00 & 0.00 & 0.00 \\
    \bottomrule
  \end{tabular}
  }
  \caption{Comparison of multi-view 3D occupancy prediction for the 16 longest tail categories (bottom 20\%). Tail-mIoU is the mean of long tail categories listed in the table. The best results are \textbf{bold}. $^*$: reproduced results; $^\dagger$: adapted to indoor scenes; ``ee.'': ``excercise equipment''.}
  \label{tab:result_longest_tail_categories}
\end{table*}


\section{Experiment}
\subsection{Experimental Setup}
\paragraph{Dataset} All of our experiments are conducted on the large-scale indoor 3D perception dataset, EmbodiedScan~\cite{wang2024embodiedscan}. The dataset integrates and augments RGB-D scan data from three classic real-world datasets: ScanNet~\cite{dai2017scannet}, 3RScan~\cite{wald2019rio}, and Matterport3D~\cite{chang2017matterport3d}. It comprises over 5,000 scanned scenes, each providing an image sequence composed of ego-centric (first-person) view RGB images and their corresponding depth maps. For the task of semantic occupancy prediction, EmbodiedScan provides dense, voxel-level semantic occupancy ground truth. Within the ego-centric coordinate system, the spatial range for occupancy prediction is defined as [-3.2m, 3.2m] for the X and Y axes, and [-0.78m, 1.78m] for the Z axis. This volume is discretized into a $40{\times}40{\times}16$ voxel grid. Each voxel is assigned a label from one of 81 categories, which include 80 indoor object classes and one ``empty'' class.
Following EmbodiedScan~\cite{wang2024embodiedscan}, we sample ten views of RGB-D frames for each scene, which form the multi-view RGB-D inputs to our framework.

\paragraph{Network Architecture} Our model architecture followed the multi-view RGB-D setting of the EmbodiedScan benchmark, employing an ImageNet-pretrained ResNet-50~\cite{he2016deep} as the 2D backbone and a MinkResNet-34~\cite{choy20194d} as the 3D backbone. The key innovation lies in our feature decoder, where we apply differentiated supervision to the multi-scale features from the 3D FPN~\cite{kuang2020voxel}. Specifically, for the highest-resolution ($40{\times}40{\times}16$) features, we attach a parallel ``main-expert" cluster. The cluster consists of a main head for the global 81-way classification (also used for inference) and 12 grouped expert heads that provide auxiliary supervision on semantic subgroups during training. For all mid- and low-resolution ($20{\times}20{\times}8$ and $10{\times}10{\times}4$) features, we only apply a standard 81-way classification head for deep supervision. All prediction heads are implemented as lightweight $1 \times 1 \times 1$ 3D convolutional layers.

\begin{table*}[h]
  \centering
  \resizebox{\linewidth}{!}{
  \begin{tabular}{l|c|c c c c c c c c c c c c c c c c}

    \toprule

    Method & mIoU(\%) & empty	& floor	& wall	& chair	& cabinet	& door	& table	& couch	& shelf	& window	& bed	& curtain	& refri.	& plant	& stairs & toilet	\\
    
    \midrule
EmbodiedScan* & 20.39 & 73.62 & 69.40 &54.90 & 52.56 & 28.59 & 33.80 & 39.99 & 44.91 & 40.63 & 28.05 & 46.87 & 48.53 & 15.75 & 34.35 & 35.92 & 63.54\\

EmbodiedScan* \textit{w.} Group & 21.96 & \textbf{74.07} & 70.50 & 54.35 & 53.99 & 31.88 & 33.19 & 42.49 & 48.95 & 42.72 & 27.98 & 51.97 & 47.63 & 18.11 & 37.06 & 34.98 & 64.22\\

EmbodiedScan* \textit{w.} Focal & 20.54 & 74.01	& 70.78	& 54.16	& 53.43	& 30.46	& 33.09	& 41.41	& 44.16	& 41.66	& 27.98	& 49.46	& 48.33	& 16.92	& 37.44	& 34.49	& 63.37\\

EmbodiedScan* \textit{w.} Dice & 20.38 & 73.35	& 69.27	& \textbf{56.22}	& 53.10	& 31.12	& 34.72	& 42.60	& 52.08	& 44.34	& 29.19	& 57.46	& 53.70	& 10.76	& 36.77	& 39.46	& \textbf{67.27}\\

EmbodiedScan* \textit{w.} UFD & 21.24 & 73.27 &70.25 & 54.01 & 54.27 & 31.89 & 33.19 & 44.07 & 51.56 & 44.24 & 29.29 & 56.84 & 52.73 & 20.25 & 35.28 & 39.54 & 65.53\\

    \midrule
Group-UFD Occ (Ours) & \textbf{22.71} & 73.99 & \textbf{71.13} & 54.92 & \textbf{55.92} & \textbf{34.15} & \textbf{36.84} & \textbf{44.47} & \textbf{54.19} & \textbf{45.70} & \textbf{29.64} & \textbf{59.32} & \textbf{54.20} & \textbf{24.60} & \textbf{39.32} & \textbf{42.74} & 66.81\\

    \bottomrule
  \end{tabular}
  }
  \caption{Ablation study on the grouping strategy and UFD loss. The table displays the results on the 16 most common indoor categories. The best results are \textbf{bold}. ``refri.'': ``refrigerator''.} 
  \label{tab:abl_result_common_categories}
\end{table*}

\paragraph{Baselines} We compare our proposed Group-UFD Occ framework against several methods, which include 1) EmbodiedScan~\cite{wang2024embodiedscan}: A framework that processes RGB-D inputs through two parallel branches and then fuse them via a dense decoder to generate the final occupancy prediction. 2) EmbodiedScan*: Our re-implementation of the EmbodiedScan baseline, conducted to evaluate performance on tail categories not reported in the original work. It also serves as the baseline model. 3) COTR$^\dagger$~\cite{ma2024cotr}: A method that introduces a coarse-to-fine semantic grouping strategy to enhance semantic discriminability in outdoor driving scenes. We adapt its grouping concept to the indoor domain for comparison.

\begin{figure*}[!t]  
    \centering
    
    \includegraphics[width=0.95\textwidth, keepaspectratio]{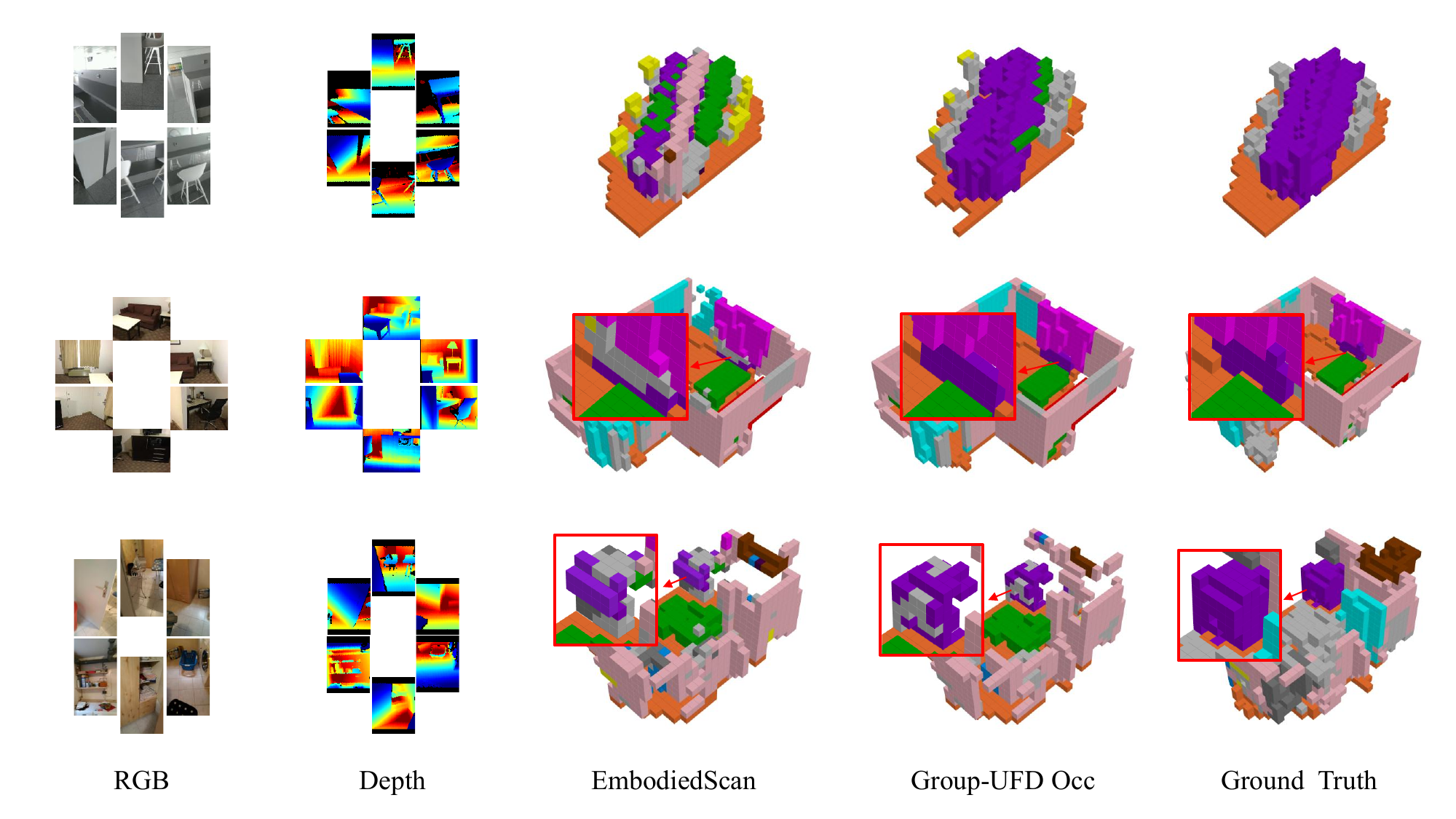}
    \caption{Qualitative comparison of EmbodiedScan and our Group-UFD Occ. For visualization clarity, we filtered out false positive predictions that occur in ground-truth empty regions.}  
    \label{f3}
\end{figure*}

\begin{table*}[htbp]
  \centering
  \resizebox{\linewidth}{!}{
  \begin{tabular}{l|c|c|c c c c c c c c c c c c c c c c}

    \toprule

    Method & mIoU(\%) & Tail-mIoU(\%) & commode	& tv & monitor & radiator & shower & mat & ottoman & wm. & towel & column & wf. & rack & stove & pillar & blinds & bar\\
    
    \midrule
EmbodiedScan* & 20.39 & 13.28 & 0.99 & 21.82 & 52.13 & 38.51 & 1.06 & 2.00 & 14.57 & 17.95 & 22.56 & 2.84 & 0.13 & 0.50 & 17.29 & \textbf{5.21}	& \textbf{1.84} & 13.04 \\

EmbodiedScan* \textit{w.} Group & 21.96 & 14.95	& 3.90	& \textbf{26.92}	& 51.67	& 42.17	& 1.10	& 1.09	& 13.73	& \textbf{31.42}	& 28.36	& \textbf{5.14}	& \textbf{0.54}	& \textbf{0.71}	& 16.90	& 3.56	& 0.54	& 11.41 \\

EmbodiedScan* \textit{w.} Focal & 20.54 & 13.58	& 1.47	& 25.37	& 51.53	& 37.32	& 0.60	& \textbf{2.42} & 11.86	& 25.25	& 24.02	& 3.69	& 0.20	& 0.35 & 16.47	& 5.14	& 0.23	& 11.37	\\

EmbodiedScan* \textit{w.} Dice & 20.38 & 14.87	& 2.46	& 18.52	& 54.28	& 42.33	& 0.75	& 0.00	& 14.50	& 21.77	& 27.51	& 0.00	& 0.00	& 0.00	& 9.90	& 0.00	& 0.00	& \textbf{45.96}\\

EmbodiedScan* \textit{w.} UFD & 21.24 & 16.23	& \textbf{4.73}	& 24.75	& \textbf{56.08}	& 42.32	& 2.88	& 0.00	& \textbf{21.09}	& 22.76	& 25.80	& 0.11	& 0.00	& 0.00	& 13.67	& 0.11	& 0.00	& 45.37	\\

    \midrule
Group-UFD Occ (Ours) & \textbf{22.71}  & \textbf{17.12} & 4.29 & 24.12 & 54.57 & \textbf{43.63} & \textbf{3.03} & 0.18	& 19.43	& 26.75	& \textbf{28.58}	& 1.20	& 0.00	& 0.00	& \textbf{22.39}	& 0.00	& 0.00 & 45.83 \\

    \bottomrule
  \end{tabular}
  }
  \caption{Ablation study on the grouping strategy and UFD loss. The table displays the results of 16 consecutive tail categories. Tail-mIoU is the mean of long tail categories list in the table. The best results are \textbf{bold}. ``refri.'': ``refrigerator'', ``wm.'': ``washing machine'', ``wf.'': ``window frame''. }
  \label{tab:abl_result_tail_categories}
\end{table*}

\begin{table*}[htbp]
  \centering
  \resizebox{\linewidth}{!}{
  \begin{tabular}{l|c|c c c c c c c c c c c c c c c c}

    \toprule

    Method & mIoU(\%) & empty	& floor	& wall	& chair	& cabinet	& door	& table	& couch	& shelf	& window	& bed	& curtain	& refri.	& plant	& stairs & toilet	\\
    
    \midrule
EmbodiedScan* & 20.39 & 73.62 & 69.40 & \textbf{54.90} & 52.56 & 28.59 & \textbf{33.80} & 39.99 & 44.91 & 40.63 & \textbf{28.05} & 46.87 & \textbf{48.53} & 15.75 & 34.35 & \textbf{35.92} & 63.54 \\

    \midrule
EmbodiedScan* \textit{w.} Group 6 & 20.45 	& 73.89	& 70.01	& 54.56	& 52.16	& 29.49	& 30.52	& 40.77	& 44.73	& 42.22	& 27.42	& 48.93	& 45.50	& \textbf{19.11}	& 36.75	& 35.51	& 60.56	\\

EmbodiedScan* \textit{w.} Group 12 & \textbf{21.96} & \textbf{74.07} & \textbf{70.50} & 54.35 & \textbf{53.99} & \textbf{31.88} & 33.19 & \textbf{42.49} & \textbf{48.95} & \textbf{42.72} & 27.98 & \textbf{51.97} & 47.63 & 18.11 & \textbf{37.06} & 34.98 & \textbf{64.22}\\

EmbodiedScan* \textit{w.} Group 17 & 18.73 & 73.74 & 70.00 & 52.83 & 52.82 & 27.43 & 29.06 & 40.71 & 44.72 & 40.88 & 25.27 & 49.40 & 44.67 & 13.48 & 35.66 & 34.87 & 57.33\\

    \bottomrule
  \end{tabular}
  }
  \caption{Ablation study on the number of semantic groups. The table displays the results on the 16 most common indoor categories. The best results are \textbf{bold}. ``refri.'': ``refrigerator''. }
  \label{tab:group_result_group_common_categories}
\end{table*}

\begin{table*}[htbp]
  \centering
\setlength{\abovecaptionskip}{3pt}
\setlength{\belowcaptionskip}{3pt}
  \resizebox{\linewidth}{!}{
  \begin{tabular}{l|c|c|c c c c c c c c c c c c c c c c}

    \toprule

    Method & mIoU(\%) & Tail-mIoU(\%) & commode	& tv & monitor & radiator & shower & mat & ottoman & wm. & towel & column & wf. & rack & stove & pillar & blinds & bar\\
    
    \midrule
EmbodiedScan* & 20.39 & 13.28 & 0.99 & 21.82 & 52.13 & 38.51 & 1.06 & \textbf{2.00} & \textbf{14.57} & 17.95 & 22.56 & 2.84 & 0.13 & 0.50 & 17.29 & \textbf{5.21}	& \textbf{1.84} & \textbf{13.04} \\					

    \midrule
EmbodiedScan* \textit{w.} Group 6 & 20.45 & 13.43	& 3.25	& 25.05	& \textbf{52.95}	& 38.46	& 0.98	& 0.91	& 10.99	& 20.49	& 22.04	& 3.15	& 0.50	& 0.23	& \textbf{18.91}	& 3.64	& 1.22	& 12.05\\																					

EmbodiedScan* \textit{w.} Group 12 & \textbf{21.96} & \textbf{14.95}	& \textbf{3.90}	& \textbf{26.92}	& 51.67	& \textbf{42.17}	& \textbf{1.10}	& 1.09	& 13.73	& \textbf{31.42}	& \textbf{28.36}	& \textbf{5.14}	& \textbf{0.54}	& \textbf{0.71}	& 16.90	& 3.56	& 0.54	& 11.41\\

EmbodiedScan* \textit{w.} Group 17 & 18.73 & 11.33	& 2.21	& 21.31	& 50.27	& 34.46	& 0.81	& 0.17	& 13.27	& 11.71	& 17.53	& 2.60	& 0.00	& 0.64	& 13.82	& 3.30	& 0.77	& 8.43\\																					

    \bottomrule
  \end{tabular}
  }
  \caption{Ablation study on the number of semantic groups. The table displays the results of 16 consecutive tail categories. Tail-mIoU is the mean of long tail categories list in the table. The best results are \textbf{bold}. ``refri.'': ``refrigerator'', ``wm.'': ``washing machine'', ``wf.'': ``window frame''. }
  \label{tab:group_result_group_tail_categories}
\end{table*}

\subsection{Experimental Results}

\paragraph{Common Categories} From the comparison results in Table~\ref{tab:main_result}, we can see that our Group-UFD Occ achieves the best performance in terms of overall mIoU. 
First, our reproduced baseline, EmbodiedScan*, achieves an mIoU of 20.39\%, slightly outperforming the original paper's reported score of 19.97\%, which validates our experimental setup. Second, we observe that COTR, which employs a coarse-to-fine grouping strategy designed for outdoor scenes, shows a slight performance degradation compared to the baseline (20.31\% vs. 20.39\% mIoU). This suggests that its grouping mechanism, which might merge distinct fine-grained indoor categories, is not well-suited for indoor semantic segmentation. In contrast, our Group-UFD Occ model boosts the mIoU to 22.71\% and demonstrates improved or superior performance on nearly all common categories, highlighting the effectiveness of our tailored grouping strategy and the UFD loss. For instance, on the \textit{door} and \textit{window} categories, COTR's performance degrades by 7.0\% and 8.4\% respectively
, whereas our Group-UFD Occ boosts performance by 9.0\% and 5.7\%.

\paragraph{Tail Categories}
Table~\ref{tab:minor_result} and Table~\ref{tab:result_longest_tail_categories} presents a performance comparison respectively on bottom 50\% and bottom 20\% tail categories among EmbodiedScan, COTR, and our Group-UFD Occ. 
The experimental results show that the method utilizing COTR's grouping strategy outperforms our reproduced EmbodiedScan baseline on tail categories. Specifically, it surpasses the EmbodiedScan baseline on 10 of the 16 bottom 50\% tail classes, and its Tail-mIoU is slightly higher. However, its mIoU results on the last 20\% of categories are worse than the baseline, indicating that this grouping method is not effective enough for indoor scene categories.
Our Group-UFD Occ achieves the highest Tail-mIoU on both bottom 50\% and bottom 20\% tail categories. For example, on the towel and washing-machine classes, our method improves upon the baseline by 26.7\% and 49.0\% respectively. Most notably, for the bar category, the improvement is a remarkable 251.4\%. Although the COTR-based method shows modest gains on some tail classes, it does so at the cost of sacrificing recognition accuracy on some common categories. 

We note that our method does not work for certain irregular and flat-shaped
rare classes. 
We analyze that this non-uniform progression stems primarily from two aspects.
First, although our semantic grouping adheres to the principle of semantic cohesion, the subsequent frequency-based blending process introduces a degree of inter-group coalescence, which partially dilutes the semantic consistency within specific groups. Furthermore, driven by the optimization objective of macro-averaged metric maximization, the localized expert prediction heads inevitably bias their optimization paths toward geometrically regular tail targets that yield higher marginal performance returns. 
Second, the structural characteristics of the UFD loss present inherent limitations when tackling these extreme cases. Although the Focal loss component down-weights frequent, easy background voxels via the focusing factor to refocus the network on hard tokens, the absolute voxel volume of geometrically complex tail classes remains too sparse.
Meanwhile, the Dice loss component, which evaluates regional intersection-over-union, exhibits an exceptionally low geometric tolerance when dealing with highly irregular or slender shapes. 
Our framework integrates multi-layer semantic grouping supervision with the UFD loss, improving performance across a broad spectrum of challenging tail categories.

\paragraph{Visualization} Figure \ref{f3} presents a qualitative comparison between our method and the baseline. The visualization results highlight the superior performance of our Group-UFD Occ. Scene 1 showcases a scene containing a bar, a tail category on which our Group-UFD Occ demonstrates a strong performance advantage. Compared to the baseline, our method more accurately recognizes and reconstructs the partition structure on top of the bar's counter. Scene 2 displays a living room environment where we focus on the tail category radiator. Our method clearly distinguishes the boundary between the fireplace and the curtain behind it, a benefit we attribute to the strong geometric integrity modeling capability provided by our UFD loss.
In Scene 3, we present a laundry room scenario. Due to the scarcity of the washing machine class in the dataset, neither model performs perfectly in reconstructing its geometry. However, our method more accurately identifies the front-loading door structure of the washing machine compared to the baseline, which sideways demonstrates the advantages of our approach.


\subsection{Ablation Study}
To validate the independent contributions of our proposed Grouping Strategy and UFD loss, we conducted detailed ablation studies in Tables~\ref{tab:abl_result_common_categories},~\ref{tab:abl_result_tail_categories},~\ref{tab:group_result_group_common_categories} and \ref{tab:group_result_group_tail_categories}. We evaluated the effect of each component on both head categories and tail categories by applying it independently to the EmbodiedScan baseline model.

\paragraph{Grouping Strategy and UFD Loss} 
We compare the following six model configurations. EmbodiedScan*: our reproduced baseline model. EmbodiedScan* w. Group: the baseline model augmented only with our 12-group hierarchical grouping strategy. EmbodiedScan* \textit{w.} Focal: the baseline model with Focal loss replace original CE loss. EmbodiedScan* \textit{w.} Dice: the baseline model with Dice loss replace original auxiliary losses. EmbodiedScan* w. UFD: the baseline model with only its loss function replaced by our UFD loss. Group-UFD Occ (Ours): our full, final method.

First, the overall performance comparison in Table~\ref{tab:abl_result_common_categories} shows that when the grouping strategy is introduced independently on top of the baseline (w. Group), the overall mIoU improves from 20.39\% to 21.96\%. This demonstrates the potential of the grouping strategy to enhance performance for general-purpose occupancy prediction. Replacing the main loss (CE loss) with Focal loss (w. Focal) slightly improves the overall mIoU, and replacing two auxiliary losses with Dice loss (w. Dice) leaves the overall mIoU almost unchanged. From Table~\ref{tab:abl_result_common_categories} and Table~\ref{tab:abl_result_tail_categories}, we can see that the introduction of Dice loss improved the IoU of most common categories and some tail classes, but also reduced the IoU of some tail classes that originally performed poorly. This is because although Dice loss's idea of directly optimizing regional intersection-over-union is well suited to the accuracy indicators of 3D occupancy prediction and effectively improves the mIoU of some tail classes, it exhibits an exceptionally low geometric tolerance when dealing with highly irregular or slender shapes. So it often tends to give up unpredictable irregular items. However, when we combine the Focal and Dice loss (w. UFD), the mIoU increases to 21.24\%, indicating that the synergistic loss function serves as an effective utility in the model optimization process. 

Second, the combined application of both components yields a synergistic effect. Our final model, Group-UFD Occ, achieves an mIoU of 22.71\%, which proves that there is a strong positive synergy between our proposed architectural innovation and the loss function optimization, leading to performance gains on all representative common classes except for empty.
Table~\ref{tab:abl_result_tail_categories} further reveals the impact of these two improvements on tail categories. Although the improvement in long tail performance is relatively small when introducing Focal loss alone and there is a performance imbalance effect when introducing Dice loss alone, it can be seen that both w. Group and w. UFD boost the Tail-mIoU, increasing it from the initial 13.28\% to 14.95\% and 16.23\%, respectively. This indicates that our method, whether at the architectural or the optimization level, can effectively shift the model's attention to more rare classes. For example, on the commode class, both w. Group (3.90\%) and w. UFD (4.73\%) far surpass the baseline's 0.99\%. Ultimately, our Group-UFD Occ model also achieves the best performance on Tail-mIoU (17.12\%), showcasing dominant advantages on several critical tail classes such as \textit{stove} 22.39\% and \textit{bar} 45.83\%.


\paragraph{Number of Semantic Groups} To explore the optimal granularity of our grouping strategy, we conduct experiments on different numbers of semantic groups. Table~\ref{tab:group_result_group_common_categories} illustrates the impact of varying group numbers on the model's overall performance.
The baseline model can be considered as having a group number of 1. The results indicate that increasing the number of groups can improve model performance to a certain extent, although the effect is limited with a small number of groups (e.g., 6 groups). The model's performance peaks when the number of groups reaches 12. It achieves an mIoU of 21.96\% and attains the best results on multiple categories such as \textit{chair}, \textit{cabinet}, \textit{couch}, \textit{bed}, and \textit{toilet}. 
However, when the grouping is further refined to 17 groups, the performance drops sharply to 18.73\%. This is likely due to the model becoming overly complex, leading to optimization difficulties from an excessive number of auxiliary tasks, and potential overfitting on some of the smaller groups. 

Similarly, Table~\ref{tab:group_result_group_tail_categories} presents the effect of different grouping granularities on the performance of long-tail categories. The same ``peaking" trend is validated here as well, with the Tail-mIoU also reaching its maximum at 12 groups. Within this configuration, the performance improvement on tail categories is the most significant overall. However, we observe that even in the worst-performing 17-group scheme, the performance on certain tail categories (e.g., \textit{commode} at 2.21\%) is still superior to the baseline. This observation sideways corroborates the general effectiveness of the grouping strategy in guiding the model's attention toward tail classes.
Therefore, we ultimately select the 12-group scheme as our optimal grouping strategy.

\section{Conclusion}
In this paper, we introduced Group-UFD Occ, a novel framework for long-tailed semantic occupancy prediction in indoor scenes. Our core contribution is a two-pronged strategy to mitigate the challenges of class imbalance. 
Extensive experiments on the EmbodiedScan dataset demonstrate that Group-UFD Occ surpasses the counterparts on both common and tail classes. We hope this work can spur further research into long-tailed occupancy prediction and accelerate its adoption in real-world applications.

Despite its success, our method has two primary limitations. First, the LLM-based semi-automated semantic grouping strategy still requires manual verification to maintain semantic consistency. Second, the introduction of UFD loss inevitably inherits the inherent limitations of Dice loss. Due to minimal regional overlap in the early stages of training, the model struggles to effectively optimize tail categories characterized by highly irregular shapes.

\section*{Acknowledgments}
During the preparation of this work, the authors used Gemini to polish the language and improve the readability of the manuscript. After using this tool, the authors reviewed and edited the content as needed and assume full responsibility for the content of the submitted manuscript.

\bibliographystyle{IEEEtran} 
\bibliography{main} 

@inproceedings{wang2024embodiedscan,
  title={Embodiedscan: A holistic multi-modal 3d perception suite towards embodied ai},
  author={Wang, Tai and Mao, Xiaohan and Zhu, Chenming and Xu, Runsen and Lyu, Ruiyuan and Li, Peisen and Chen, Xiao and Zhang, Wenwei and Chen, Kai and Xue, Tianfan and others},
  booktitle={CVPR},
  pages={19757--19767},
  year={2024}
}

@article{liu2025aligning,
  title={Aligning cyber space with physical world: A comprehensive survey on embodied ai},
  author={Liu, Yang and Chen, Weixing and Bai, Yongjie and Liang, Xiaodan and Li, Guanbin and Gao, Wen and Lin, Liang},
  journal={IEEE/ASME Transactions on Mechatronics},
  year={2025},
  publisher={IEEE}
}

@article{guan2024gramo,
  title={GRAMO: geometric resampling augmentation for monocular 3D object detection},
  author={Guan, He and Song, Chunfeng and Zhang, Zhaoxiang},
  journal={Frontiers of Computer Science},
  volume={18},
  number={5},
  pages={185706},
  year={2024},
  publisher={Springer}
}

@inproceedings{tong2023scene,
  title={Scene as occupancy},
  author={Tong, Wenwen and Sima, Chonghao and Wang, Tai and Chen, Li and Wu, Silei and Deng, Hanming and Gu, Yi and Lu, Lewei and Luo, Ping and Lin, Dahua and others},
  booktitle={ICCV},
  pages={8406--8415},
  year={2023}
}

@inproceedings{ma2024cotr,
  title={Cotr: Compact occupancy transformer for vision-based 3d occupancy prediction},
  author={Ma, Qihang and Tan, Xin and Qu, Yanyun and Ma, Lizhuang and Zhang, Zhizhong and Xie, Yuan},
  booktitle={CVPR},
  pages={19936--19945},
  year={2024}
}

@inproceedings{tang2024sparseocc,
  title={Sparseocc: Rethinking sparse latent representation for vision-based semantic occupancy prediction},
  author={Tang, Pin and Wang, Zhongdao and Wang, Guoqing and Zheng, Jilai and Ren, Xiangxuan and Feng, Bailan and Ma, Chao},
  booktitle={CVPR},
  pages={15035--15044},
  year={2024}
}

@article{xu2025survey,
  title={A survey on occupancy perception for autonomous driving: The information fusion perspective},
  author={Xu, Huaiyuan and Chen, Junliang and Meng, Shiyu and Wang, Yi and Chau, Lap-Pui},
  journal={Information Fusion},
  volume={114},
  pages={102671},
  year={2025},
  publisher={Elsevier}
}

@article{zhang2026vision,
  title={Vision-based 3d occupancy prediction in autonomous driving: a review and outlook},
  author={Zhang, Yanan and Zhang, Jinqing and Wang, Zengran and Xu, Junhao and Huang, Di},
  journal={Frontiers of Computer Science},
  volume={20},
  number={1},
  pages={2001301},
  year={2026},
  publisher={Springer}
}

@inproceedings{milletari2016v,
  title={V-net: Fully convolutional neural networks for volumetric medical image segmentation},
  author={Milletari, Fausto and Navab, Nassir and Ahmadi, Seyed-Ahmad},
  booktitle={3DV},
  pages={565--571},
  year={2016},
  organization={Ieee}
}

@inproceedings{lin2017focal,
  title={Focal loss for dense object detection},
  author={Lin, Tsung-Yi and Goyal, Priya and Girshick, Ross and He, Kaiming and Doll{\'a}r, Piotr},
  booktitle={ICCV},
  pages={2980--2988},
  year={2017}
}

@inproceedings{dai2017scannet,
  title={Scannet: Richly-annotated 3d reconstructions of indoor scenes},
  author={Dai, Angela and Chang, Angel X and Savva, Manolis and Halber, Maciej and Funkhouser, Thomas and Nie{\ss}ner, Matthias},
  booktitle={CVPR},
  pages={5828--5839},
  year={2017}
}

@inproceedings{wald2019rio,
  title={Rio: 3d object instance re-localization in changing indoor environments},
  author={Wald, Johanna and Avetisyan, Armen and Navab, Nassir and Tombari, Federico and Nie{\ss}ner, Matthias},
  booktitle={ICCV},
  pages={7658--7667},
  year={2019}
}

@inproceedings{chang2017matterport3d,
  title={Matterport3D: Learning from RGB-D Data in Indoor Environments},
  author={Chang, Angel and Dai, Angela and Funkhouser, Thomas and Halber, Maciej and Niebner, Matthias and Savva, Manolis and Song, Shuran and Zeng, Andy and Zhang, Yinda},
  booktitle={3DV},
  pages={667--676},
  year={2017},
  organization={IEEE}
}

@inproceedings{silberman2012indoor,
  title={Indoor segmentation and support inference from rgbd images},
  author={Silberman, Nathan and Hoiem, Derek and Kohli, Pushmeet and Fergus, Rob},
  booktitle={ECCV},
  pages={746--760},
  year={2012},
  organization={Springer}
}

@inproceedings{song2017semantic,
  title={Semantic scene completion from a single depth image},
  author={Song, Shuran and Yu, Fisher and Zeng, Andy and Chang, Angel X and Savva, Manolis and Funkhouser, Thomas},
  booktitle={CVPR},
  pages={1746--1754},
  year={2017}
}

@inproceedings{cao2022monoscene,
  title={Monoscene: Monocular 3d semantic scene completion},
  author={Cao, Anh-Quan and De Charette, Raoul},
  booktitle={CVPR},
  pages={3991--4001},
  year={2022}
}

@inproceedings{huang2023tri,
  title={Tri-perspective view for vision-based 3d semantic occupancy prediction},
  author={Huang, Yuanhui and Zheng, Wenzhao and Zhang, Yunpeng and Zhou, Jie and Lu, Jiwen},
  booktitle={CVPR},
  pages={9223--9232},
  year={2023}
}

@inproceedings{yu2024monocular,
  title={Monocular occupancy prediction for scalable indoor scenes},
  author={Yu, Hongxiao and Wang, Yuqi and Chen, Yuntao and Zhang, Zhaoxiang},
  booktitle={ECCV},
  pages={38--54},
  year={2024},
  organization={Springer}
}

@inproceedings{li2025sliceocc,
  title={SliceOcc: Indoor 3D semantic occupancy prediction with vertical slice representation},
  author={Li, Jianing and Lu, Ming and Liu, Juntao and Wang, Hao and Gu, Chenyang and Zheng, Wenzhao and Du, Li and Zhang, Shanghang},
  booktitle={ICRA},
  pages={15762--15768},
  year={2025},
  organization={IEEE}
}

@inproceedings{pan2024renderocc,
  title={Renderocc: Vision-centric 3d occupancy prediction with 2d rendering supervision},
  author={Pan, Mingjie and Liu, Jiaming and Zhang, Renrui and Huang, Peixiang and Li, Xiaoqi and Xie, Hongwei and Wang, Bing and Liu, Li and Zhang, Shanghang},
  booktitle={ICRA},
  pages={12404--12411},
  year={2024},
  organization={IEEE}
}

@inproceedings{wu2025embodiedocc,
  title={Embodiedocc: Embodied 3d occupancy prediction for vision-based online scene understanding},
  author={Wu, Yuqi and Zheng, Wenzhao and Zuo, Sicheng and Huang, Yuanhui and Zhou, Jie and Lu, Jiwen},
  booktitle={ICCV},
  pages={26360--26370},
  year={2025}
}

@inproceedings{liu2021lmscnet,
  title={LMSCNet: Lightweight Multi-scale 3D Semantic Completion},
  author={Liu, Yan-Ting and Chiu, Hsuan-Chao and Chen, Yu-An and Hsu, Chi-Che and Chen, Wei-Chao and Chen, Bing-Yu},
  booktitle={3DV},
  pages={856--865},
  year={2021},
  organization={IEEE}
}

@inproceedings{wang2023openoccupancy,
  title={Openoccupancy: A large scale benchmark for surrounding semantic occupancy perception},
  author={Wang, Xiaofeng and Zhu, Zheng and Xu, Wenbo and Zhang, Yunpeng and Wei, Yi and Chi, Xu and Ye, Yun and Du, Dalong and Lu, Jiwen and Wang, Xingang},
  booktitle={ICCV},
  pages={17850--17859},
  year={2023}
}

@article{ming2024occfusion,
  title={Occfusion: Multi-sensor fusion framework for 3d semantic occupancy prediction},
  author={Ming, Zhenxing and Berrio, Julie Stephany and Shan, Mao and Worrall, Stewart},
  journal={IEEE Transactions on Intelligent Vehicles},
  year={2024},
  publisher={IEEE}
}

@inproceedings{philion2020lift,
  title={Lift, splat, shoot: Encoding images from arbitrary camera rigs by implicitly unprojecting to 3d},
  author={Philion, Jonah and Fidler, Sanja},
  booktitle={ECCV},
  pages={194--210},
  year={2020},
  organization={Springer}
}

@inproceedings{choy20194d,
  title={4d spatio-temporal convnets: Minkowski convolutional neural networks},
  author={Choy, Christopher and Gwak, JunYoung and Savarese, Silvio},
  booktitle={CVPR},
  pages={3075--3084},
  year={2019}
}

@article{kuang2020voxel,
  title={Voxel-FPN: Multi-scale voxel feature aggregation for 3D object detection from LIDAR point clouds},
  author={Kuang, Hongwu and Wang, Bei and An, Jianping and Zhang, Ming and Zhang, Zehan},
  journal={Sensors},
  volume={20},
  number={3},
  pages={704},
  year={2020},
  publisher={MDPI}
}

@inproceedings{he2016deep,
  title={Deep residual learning for image recognition},
  author={He, Kaiming and Zhang, Xiangyu and Ren, Shaoqing and Sun, Jian},
  booktitle={CVPR},
  pages={770--778},
  year={2016}
}

@article{lynam2025augmented,
  title={Augmented reality navigation: A survey},
  author={Lynam, Hudson and Dascalu, Sergiu and Folmer, Eelke},
  journal={International Journal of Human--Computer Interaction},
  volume={41},
  number={16},
  pages={10190--10206},
  year={2025},
  publisher={Taylor \& Francis}
}

@ARTICLE{11123690,
  author={Chen, Mingcong and Fan, Siqi and Cao, Guanglin and Liu, Yun-hui and Liu, Hongbin},
  journal={IEEE RA-L}, 
  title={USPilot: An Embodied Robotic Assistant Ultrasound System With a Large Language Model Enhanced Graph Planner}, 
  year={2025},
  volume={10},
  number={10},
  pages={10027-10034},
  doi={10.1109/LRA.2025.3598625}}

@ARTICLE{10740797,
  author={Zhang, Dongkun and Liang, Jiaming and Lu, Sha and Guo, Ke and Wang, Qi and Xiong, Rong and Miao, Zhenwei and Wang, Yue},
  journal={IEEE RA-L}, 
  title={PEP: Policy-Embedded Trajectory Planning for Autonomous Driving}, 
  year={2024},
  volume={9},
  number={12},
  pages={11361-11368},
  doi={10.1109/LRA.2024.3490377}}

@ARTICLE{11123741,
  author={Zhang, Sheng and Huai, Lian and Liu, Yuyu and Jiang, Xingqun},
  journal={IEEE RA-L}, 
  title={CMG3D: Compensation Towards Modality Gap for Open-Vocabulary Indoor 3D Object Detection}, 
  year={2025},
  volume={10},
  number={10},
  pages={10019-10026},
  doi={10.1109/LRA.2025.3598669}}

@ARTICLE{10325591,
  author={Song, Xiaogang and Zhou, Zhenhua and Zhang, Lei and Lu, Xiaofeng and Hei, Xinhong},
  journal={IEEE RA-L}, 
  title={PSNS-SSD: Pixel-Level Suppressed Nonsalient Semantic and Multicoupled Channel Enhancement Attention for 3D Object Detection}, 
  year={2024},
  volume={9},
  number={1},
  pages={603-610},
  doi={10.1109/LRA.2023.3335773}}

@ARTICLE{10791908,
  author={Li, Zhiqi and Wang, Wenhai and Li, Hongyang and Xie, Enze and Sima, Chonghao and Lu, Tong and Yu, Qiao and Dai, Jifeng},
  journal={IEEE T-PAMI}, 
  title={BEVFormer: Learning Bird’s-Eye-View Representation From LiDAR-Camera via Spatiotemporal Transformers}, 
  year={2025},
  volume={47},
  number={3},
  pages={2020-2036},
  doi={10.1109/TPAMI.2024.3515454}}

@article{TIAN2026131305,
title = {MambaOcc: Visual state space models for BEV-based occupancy prediction with local adaptive reordering},
journal = {Expert Systems with Applications},
volume = {310},
pages = {131305},
year = {2026},
issn = {0957-4174},
doi = {https://doi.org/10.1016/j.eswa.2026.131305},
author = {Yonglin Tian and Songlin Bai and Zhiyao Luo and Yutong Wang and Hui Zhang and Baoqing Guo and Yisheng Lv and Fei-Yue Wang}
}

\end{document}